\documentclass{article} % For LaTeX2e
\usepackage{iclr2027_conference,times}
\iclrfinalcopy

\usepackage{amsmath,amsfonts,bm}

\def\eqref#1{equation~\ref{#1}}
\def\1{\bm{1}}

\DeclareMathAlphabet{\mathsfit}{\encodingdefault}{\sfdefault}{m}{sl}
\SetMathAlphabet{\mathsfit}{bold}{\encodingdefault}{\sfdefault}{bx}{n}

\usepackage{hyperref}
\usepackage{url}
\usepackage{booktabs}

\usepackage{graphicx}
\usepackage{multirow}

\title{BiNar: Bitstream Native Restorer for Bitstream-corrupted Image Restoration}

\usepackage{capt-of}

\author{Fangcheng Li \quad Zhen Yu \quad Kejun Wu \quad Tianyi Liu}

\begin{document}

\maketitle

\begin{abstract}
Bitstream-corrupted image restoration (BCIR) aims to reconstruct faithful images from corrupted compressed bitstreams. However, corruption occurs in bitstream-domain may lead to unpredictable error propagation and distortion amplification in pixel-domain, and even result in complete undecodability by standard decoders, posing significant challenge to existing image restoration paradigms.
To address this challenge, we propose \textbf{Bi}tstream \textbf{Na}tive \textbf{R}estorer (\textbf{BiNar}), a generative restoration framework that leverages fine-grained bitstream-native semantics to guide bitstream-corrupted image restoration. 
In BiNar, Bitstream Semantic Modeling (\textbf{BSeM}) extracts robust informative semantics and models sequential bytes directly from corrupted bitstream, enabling to generate fine-grained bitstream descriptions without relying on pixel decoding.   
Based on the descriptions, the Corruption-adaptive Multimodal Semantic Aggregation (\textbf{CMSA}) further extracts adaptive visual modal cues in decoded corrupted images and aggregates to bitstream descriptions. The multimodal aggregated fine-grained descriptions can serve as coherent restoration guidance.
Finally, Blind Corruption Localization (\textbf{BCoL}) exploits the segmentation prior of visual foundation models to identify and locate error regions that requires recovery in corrupted images. 
The located error regions and the fine-grained descriptions jointly serve as visual prompt and language prompt of diffusion models image restoration, respectively.
Extensive experiments under diverse bitstream corruption conditions show that BiNar outperforms all State-of-the-arts image restoration models regardless of perceptual quality or objective quality.
Ablation studies demonstrate the effectiveness of fine-grained bitstream-native semantics in BCIR task.  
\end{abstract}

\section{Introduction}

Image restoration aims to recover clean images from degraded inputs by
exploiting the texture, structure, and semantic cues that remain visible.
Faithful restoration therefore depends on whether these cues remain
informative about the original image.
However, bitstream corruption can severely distort or even eliminate these
cues. Compressed image bitstreams are highly vulnerable to corruption during
storage and transmission. Since bitstream corruption occurs before image decoding, localized errors may
propagate through the decoding process and disrupt subsequent coded content
\cite{lam1995error,bourdon2005multi}. As a result, even limited corruption can
produce widespread and unpredictable pixel-domain distortions, including color
shifts, structural displacement, missing content, and severe visual artifacts
\cite{liu2023bitstream}. Under such degradation, semantic
modeling from the decoded image becomes increasingly challenging, as the
observed pixels may no longer faithfully represent the original content. In more severe cases, bitstream corruption can even cause decoding failure,
leaving no usable visual information from which semantic cues can be
extracted. Such degraded or missing semantics make faithful image restoration
particularly challenging.

\begin{figure*}[!htbp]
    \centering
    \includegraphics[width=\textwidth]{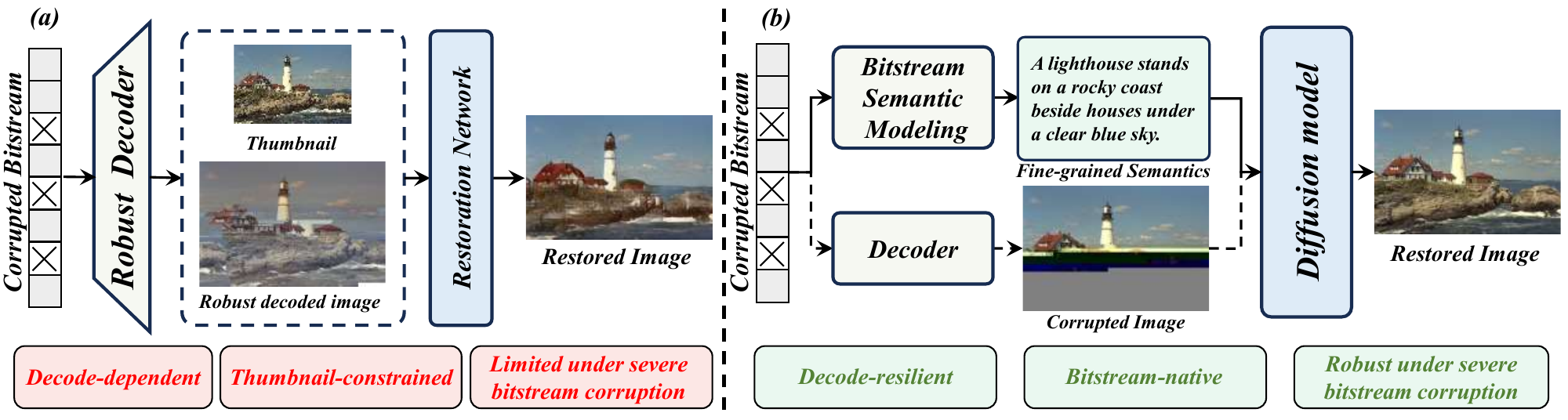}
    \caption{Paradigm comparison of bitstream-corrupted image restoration.
(a) Conventional bitstream-corrupted image restoration relies on decoded visual observations and auxiliary thumbnail guidance~\cite{liu2023bitstream}.
(b) Our method removes this dependence by extracting semantic guidance
directly from corrupted bitstream for robust image restoration.
}
    \label{fig:pipeline_comparison}
\end{figure*}

As illustrated in Fig.~\ref{fig:pipeline_comparison}, existing methods for
bitstream-corrupted image restoration (BCIR) mainly attempt to recover visual
information after decoding or exploit thumbnail guidance for restoration
\cite{liu2023bitstream,hu2025restoration}. When decoding produces severe color
distortion, structural displacement, or missing regions, the remaining pixels
provide limited information for recovering fine structures and semantic
details.
Moreover, thumbnail guidance is not universally available. Some compressed
image files do not contain an embedded thumbnail, since common image formats
such as Exif and JFIF do not require one \cite{hamilton2004jpeg}. Even when a
thumbnail is originally present, corruption of the file structure or associated
metadata may prevent it from being correctly parsed and extracted
\cite{wiegand2003overview}. As a result, this additional visual cue may also be
unavailable in challenging corruption cases, further limiting restoration
based on decoded visual information.
A bitstream error can disrupt the decoding process and propagate to subsequent content,
causing a much larger region to become distorted in the pixel ~\cite{kuo2019long}. However,
this propagation does not imply that the corresponding information has been
equally lost from the compressed representation. Large portions of the coded
data may remain unchanged even when the decoded image exhibits severe color
distortion, structural displacement, or missing regions. These surviving bytes
can therefore preserve information about the original image that is no longer
faithfully expressed in the decoded observation. This makes the corrupted
bitstream itself a promising source of semantic information for restoration.

To this end, we propose \textbf{Bitstream Native Restorer (BiNar)}, a
generative restoration framework that exploits fine-grained bitstream-native
semantics for bitstream-corrupted image restoration. The key contribution of
BiNar is to establish a semantic pathway directly from corrupted bitstreams,
providing informative restoration cues when pixel-derived semantics become
unreliable or unavailable. Specifically, \textbf{Bitstream Semantic Modeling
(BSeM)} directly models sequential bytes from corrupted bitstreams to extract
fine-grained semantics without relying on pixel decoding.
\textbf{Corruption-adaptive Multimodal Semantic Aggregation (CMSA)} further
extracts useful visual cues from decoded corrupted images and aggregates them
with the bitstream-derived semantics to obtain enhanced fine-grained
descriptions. In parallel, \textbf{Blind Corruption Localization (BCoL)}
exploits the segmentation prior of a visual foundation model to identify
regions requiring recovery. The resulting fine-grained descriptions and
localized corruption regions serve as language and visual prompts to jointly
guide a pretrained diffusion model for image restoration. Extensive
experiments across diverse bitstream corruption conditions show that BiNar
achieves the best performance on seven of eight restoration metrics, while
ablation studies further demonstrate the effectiveness of fine-grained
bitstream-native semantics and the proposed components.

\section{Related Work}
\subsection{Bitstream-Corrupted Visual Restoration}

Early restoration of corrupted compressed images mainly relied on error
concealment, using surviving spatial or transform-domain information to
reconstruct damaged content
\citep{gormish2003icon,lee2003error,bourdon2005multi}. More recent BCIR
methods improve decoding robustness and exploit auxiliary information such as
embedded thumbnails or neighboring coefficients for restoration
\citep{liu2023bitstream,hu2025restoration,jiang2025optimizing}. Generative
approaches further introduce corruption localization, coding information, and
visual foundation models to guide image or video recovery
\citep{wang2025blind,liu2023bitstream2,liu2025towards,liu2026beyond}.
However, these methods still rely heavily on decoded visual information or
auxiliary cues from the corrupted file. Under severe corruption, decoding
artifacts or decoding failure can substantially reduce the information
available for restoration, while auxiliary cues such as embedded thumbnails
may also be unavailable or inaccessible.

\subsection{Semantic-Guided Image Restoration}

Generative priors have substantially improved image restoration by enabling
plausible reconstruction beyond the information directly available in degraded
pixels. Early inpainting and diffusion-based methods recover missing content
from known image regions
\citep{yu2018generative,suvorov2022resolution,lugmayr2022repaint,rombach2022high}.
Recent restoration methods further incorporate large generative models and
semantic conditions to improve perceptual quality and content consistency
\citep{lin2024diffbir,yu2024scaling}. More explicit semantic guidance has been
introduced through vision-language models, textual prompts, and global or local
visual representations
\citep{luo2024controlling,wu2024seesr,kong2025dual}. However, such guidance is
typically derived from degraded images and may become unreliable under severe
bitstream corruption. This motivates extracting semantic information directly
from the corrupted bitstream.

\subsection{Bitstream-Domain Understanding}

Bitstream-domain understanding aims to extract visual information directly
from encoded data without relying on full pixel reconstruction. Early studies
mainly exploited structured representations available in valid compressed
images for recognition and segmentation
\citep{duan2023unified,ji2024compressed}. More recent work moves beyond
codec-specific features and models raw byte sequences directly. ByteFormer
\citep{horton2023bytes}, MEGABYTE \citep{yu2023megabyte}, MambaByte
\citep{wang2401mambabyte}, and bGPT \citep{wu2024beyond} demonstrate that
meaningful representations can be learned directly from byte-level inputs,
reducing dependence on conventional decoding.
Recent studies further extend this direction to corrupted bitstreams, where
semantic information must be recovered from incomplete or damaged encoded
data. Cibic \citep{liang2026cibic} explores semantic extraction and caption
generation from corrupted image bitstreams, while BLMSP
\citep{huang2026towards} investigates semantic priors from corrupted video
bytes for multiple downstream tasks. CBSU-ALLM \citep{wu2026corrupted}
further treats corrupted image bitstream fragments as a byte-language
modality for semantic understanding without successful pixel decoding.
Together, these studies show that corrupted bitstreams can retain useful
semantic information beyond what is recoverable from decoded pixels.

\begin{figure}[t]
    \centering
    \includegraphics[width=\linewidth]{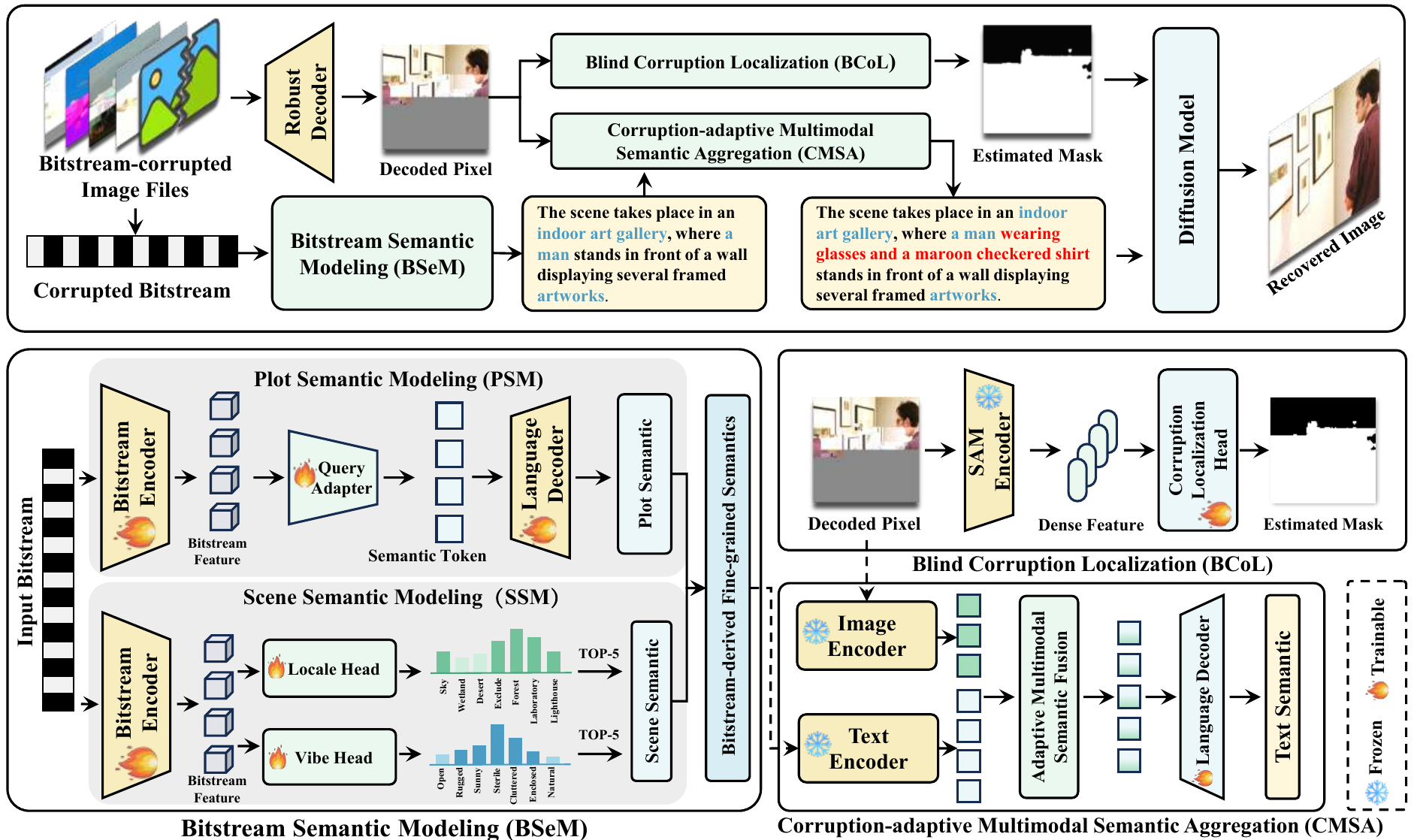}
    \caption{
Overview of \textbf{BiNar}. BSeM extracts fine-grained bitstream-native
semantics, CMSA aggregates complementary visual cues, and BCoL localizes
corrupted regions. The resulting semantic and spatial prompts jointly guide
diffusion-based restoration.
}
    \label{fig:framework}
\end{figure}
% Plot 
% scene locale vibe 
% fine-grained

\section{Methodology}
\label{sec:method}

\subsection{Framework Overview}
\label{sec:framework_overview}

We propose \textbf{Bitstream Native Restorer (BiNar)} for
bitstream-corrupted image restoration (BCIR). BiNar exploits fine-grained
bitstream-native semantics together with available visual information and
corruption localization to guide generative restoration.
As shown in Fig.~\ref{fig:framework}, \textbf{Bitstream Semantic Modeling
(BSeM)} directly models the corrupted byte sequence to extract Plot and Scene
semantics, where Scene is characterized by Locale and Vibe.
\textbf{Corruption-adaptive Multimodal Semantic Aggregation (CMSA)}
aggregates these bitstream-native semantics with useful visual cues from the
decoded corrupted image to produce a fine-grained description. In parallel,
\textbf{Blind Corruption Localization (BCoL)} identifies regions requiring
recovery. The resulting semantic description and corruption regions jointly
guide a pretrained diffusion model for image restoration.

\subsection{Bitstream Semantic Modeling}
\label{sec:bsem}

BSeM models the corrupted byte sequence from two complementary semantic
perspectives, \textbf{Plot} and \textbf{Scene}. Plot captures the main
subjects, their appearance, actions, and interactions, while Scene describes
the broader context through \textbf{Locale} and \textbf{Vibe}. Together,
they provide fine-grained bitstream-native semantics for restoration.

\paragraph{Plot Semantic Modeling (PSM).}
Given a corrupted bitstream $\tilde{\mathbf{b}}$, PSM employs a ByteFormer
encoder $E_p$ to obtain its bitstream representation. Since the resulting
representation is variable in length, a Query Adapter $A$ with a fixed set of
learnable queries aggregates it into compact semantic tokens,
\begin{equation}
    \mathbf{Z}_p = A\big(E_p(\tilde{\mathbf{b}})\big).
\end{equation}
The semantic tokens are projected into the embedding space of a language
decoder and used to condition autoregressive generation. The decoder produces
a Plot description $\mathbf{c}_p$ containing detailed subject, appearance,
action, and interaction information.

\paragraph{Scene Semantic Modeling (SSM).}
SSM models the broader scene context through \textbf{Locale} and
\textbf{Vibe}. Locale describes the type of place or scene, while Vibe
characterizes its overall environmental attributes. Two independently trained
ByteFormer-based classifiers estimate their category distributions,
\begin{equation}
    \mathbf{p}_l = C_l\big(E_l(\tilde{\mathbf{b}})\big), \qquad
    \mathbf{p}_v = C_v\big(E_v(\tilde{\mathbf{b}})\big),
\end{equation}
where $E_l$ and $E_v$ are the corresponding bitstream encoders, and
$C_l$ and $C_v$ denote the classification heads.

Rather than retaining only the Top-$1$ prediction, SSM preserves the
Top-$K$ Locale and Vibe candidates together with their confidence scores,
denoted by $\mathcal{C}_l$ and $\mathcal{C}_v$. This retains multiple
plausible scene hypotheses under corruption for subsequent aggregation.

\paragraph{Training Objectives.}
PSM is optimized with an autoregressive generation loss
$\mathcal{L}_{\mathrm{cap}}$, a token-level alignment loss
$\mathcal{L}_{\mathrm{align}}$, and a batch-wise contrastive loss
$\mathcal{L}_{\mathrm{con}}$,
\begin{equation}
    \mathcal{L}_{\mathrm{PSM}}
    =
    \mathcal{L}_{\mathrm{cap}}
    + \lambda_{\mathrm{align}}\mathcal{L}_{\mathrm{align}}
    + \lambda_{\mathrm{con}}\mathcal{L}_{\mathrm{con}}.
\end{equation}
In the first training stage, the Query Adapter and the final ByteFormer blocks
are optimized while the language decoder remains frozen. In the second stage,
ByteFormer is frozen, while the Query Adapter and low-rank adapters in the
language decoder are jointly optimized. The same objectives are used in both
stages. SSM is trained separately for Locale and Vibe prediction using
classification supervision.

The Plot description $\mathbf{c}_p$, Locale candidates
$\mathcal{C}_l$, and Vibe candidates $\mathcal{C}_v$ are subsequently
passed to CMSA.

\subsection{Corruption-adaptive Multimodal Semantic Aggregation}
\label{sec:CMSA}

CMSA aggregates the bitstream-native semantics produced by BSeM with useful
visual cues from the decoded corrupted image. Given the Plot description
$\mathbf{c}_p$, Locale candidates $\mathcal{C}_l$, Vibe candidates
$\mathcal{C}_v$, and the decoded image $\tilde{\mathbf{x}}$, the resulting
fine-grained description is formulated as
\begin{equation}
    \mathbf{c}_m =
    F_{\mathrm{CMSA}}
    \left(
        \mathbf{c}_p,
        \mathcal{C}_l,
        \mathcal{C}_v,
        \mathcal{V}(\tilde{\mathbf{x}})
    \right),
\end{equation}
where $\mathcal{V}(\tilde{\mathbf{x}})$ denotes the available visual cues
and $\mathbf{c}_m$ denotes the aggregated multimodal description.

When recognizable content remains in the decoded image, CMSA uses visual cues
to complement and refine the bitstream semantics. As decoding artifacts become
more severe, CMSA relies increasingly on the information extracted directly
from the bitstream. When decoding fails, $\mathbf{c}_m$ is constructed solely
from the BSeM outputs. The resulting description serves as the language prompt
for restoration.

\subsection{Blind Corruption Localization and Semantically Guided Restoration}
\label{sec:restoration}

BCoL identifies corrupted regions while preserving content that remains
usable in the decoded image. Given $\tilde{\mathbf{x}}$, a frozen visual
encoder $E_{\mathrm{vis}}$ extracts dense features, which are processed by a
lightweight localization head $H_c$ to predict the corruption mask
\begin{equation}
    \hat{\mathbf{m}}
    =
    \sigma\left(
        H_c\big(E_{\mathrm{vis}}(\tilde{\mathbf{x}})\big)
    \right).
\end{equation}
BCoL requires no point, box, or mask prompts. During training, the localization
head is supervised by the ground-truth corruption mask $\mathbf{m}$ using
\begin{equation}
    \mathcal{L}_{\mathrm{BCoL}}
    =
    \mathcal{L}_{\mathrm{BCE}}
    +
    \mathcal{L}_{\mathrm{Dice}},
\end{equation}
while the visual encoder remains frozen.

The predicted mask and the fine-grained description produced by CMSA provide
spatial and language guidance for a pretrained diffusion model,
\begin{equation}
    \hat{\mathbf{x}}
    =
    G\left(
        \tilde{\mathbf{x}},
        \hat{\mathbf{m}},
        \mathbf{c}_m
    \right).
\end{equation}
Here, $\hat{\mathbf{m}}$ specifies where recovery is required, while
$\mathbf{c}_m$ provides semantic guidance for reconstruction. When decoding
fails and no usable image is available, restoration falls back to the full
image region using the semantics extracted by BSeM.

\section{Experiment}

\subsection{Experimental Setup}
\label{sec:experimental_setup}

\paragraph{Dataset and Corruption.}
We construct the training and evaluation sets based on ADE20K, which contains
diverse visual scenes with annotations corresponding to the \textbf{Locale}
and \textbf{Vibe} semantics modeled in BSM. Each image is resized to
$256\times256$, center-cropped to $224\times224$, and encoded into an image
bitstream. We consider two corruption types, Bit-Flip and Byte-Loss, which
randomly invert bits and remove byte information, respectively. For each
corruption type, corruption is applied within four stream ranges, covering
$25\%$, $50\%$, $75\%$, and $100\%$ of the bitstream. Each of the resulting
eight corruption configurations is independently applied to the complete test
split, producing eight corrupted test subsets. For evaluation, each of the eight corruption configurations is applied to the
complete test split, producing a separate corrupted test subset for each
setting. All methods are evaluated on the same corrupted inputs.
\paragraph{Implementation Details.}
For semantic extraction, BiNar directly takes the corrupted bitstream as input,
while vision-language baselines operate on the corresponding decoded image.
For restoration comparison, the restoration backbone and sampling settings are
kept consistent whenever applicable. Clean images are used only as references
for training or evaluation and are never available during inference.
Additional training details are provided in the supplementary material.

\paragraph{Evaluation Metrics.}
For semantic extraction, we report BLEU-1, BLEU-4, ROUGE-L, CIDEr, Token-F1, and BERTScore-F1. For image restoration, we report eight complementary metrics, divided into objective and perceptual categories. The objective metrics include PSNR, SSIM, MAE, and VFID, which evaluate pixel-level fidelity, structural consistency, reconstruction error, and distributional similarity, respectively. The perceptual metrics include DreamSim, LPIPS, CLIP-I, and DINO-I, which assess high-level visual similarity, perceptual feature consistency, and semantic or representation-level alignment between restored and reference images.

\subsection{Experimental Results}

Since fine-grained semantic descriptions serve as the language guidance for
restoration in BiNar, their quality directly affects the fidelity of the
reconstructed content. We therefore first evaluate whether BiNar can obtain
more accurate and robust semantics under bitstream corruption, and further
examine whether this semantic advantage translates into better restoration
quality. We then compare the complete BiNar framework with existing image
restoration methods.

\subsubsection{Evaluations on  Bitstream-native Semantics}
\label{sec:semantic_effectiveness}

\begin{figure*}[!t]
  \centering

  % =========================================================
  % Quantitative Semantic Comparison
  % =========================================================
  \captionof{table}{Quantitative comparison of fine-grained semantic
  descriptions under bitstream corruption. Higher is better for all metrics.}
  \label{tab:caption_comparison}

  \scriptsize
  \resizebox{\textwidth}{!}{%
  \begin{tabular}{lccccccc}
    \toprule
    Method & BLEU-1 & BLEU-4 & ROUGE-L & CIDEr & Token-F1
      & Micro-F1 & BERTScore-F1 \\
    \midrule

    GLM-4.6V
      & 0.1797 & 0.0505 & 0.1917 & 0.2707
      & 0.2774 & 0.1969 & 0.6329 \\

    InternVL3
      & 0.1065 & 0.0242 & 0.1649 & 0.0958
      & 0.2411 & 0.1555 & 0.6323 \\

    Qwen3-VL
      & 0.0243 & 0.0061 & 0.1512 & 0.0236
      & 0.2141 & 0.1405 & 0.6330 \\

    GPT-5.6
      & 0.2794 & 0.0693 & 0.1953 & 0.1440
      & 0.3463 & 0.1939 & 0.9719 \\

    Gemini-3.8-Flash
      & 0.0677 & 0.0169 & 0.1201 & 0.0186
      & 0.1932 & 0.1388 & 0.9705 \\

    Claude Opus 4.6
      & 0.2593 & 0.0648 & 0.2093 & 0.1627
      & 0.3130 & 0.1530 & 0.9698 \\

    \textbf{BiNar (Ours)}
      & \textbf{0.4541}
      & \textbf{0.2248}
      & \textbf{0.4367}
      & \textbf{0.2934}
      & \textbf{0.5028}
      & \textbf{0.3435}
      & \textbf{0.9776} \\

    \bottomrule
  \end{tabular}%
  }

  \vspace{1.5mm}

  % =========================================================
  % Qualitative Semantic Comparison
  % =========================================================
  \includegraphics[
      width=\linewidth,
      height=0.38\textheight,
      keepaspectratio
  ]{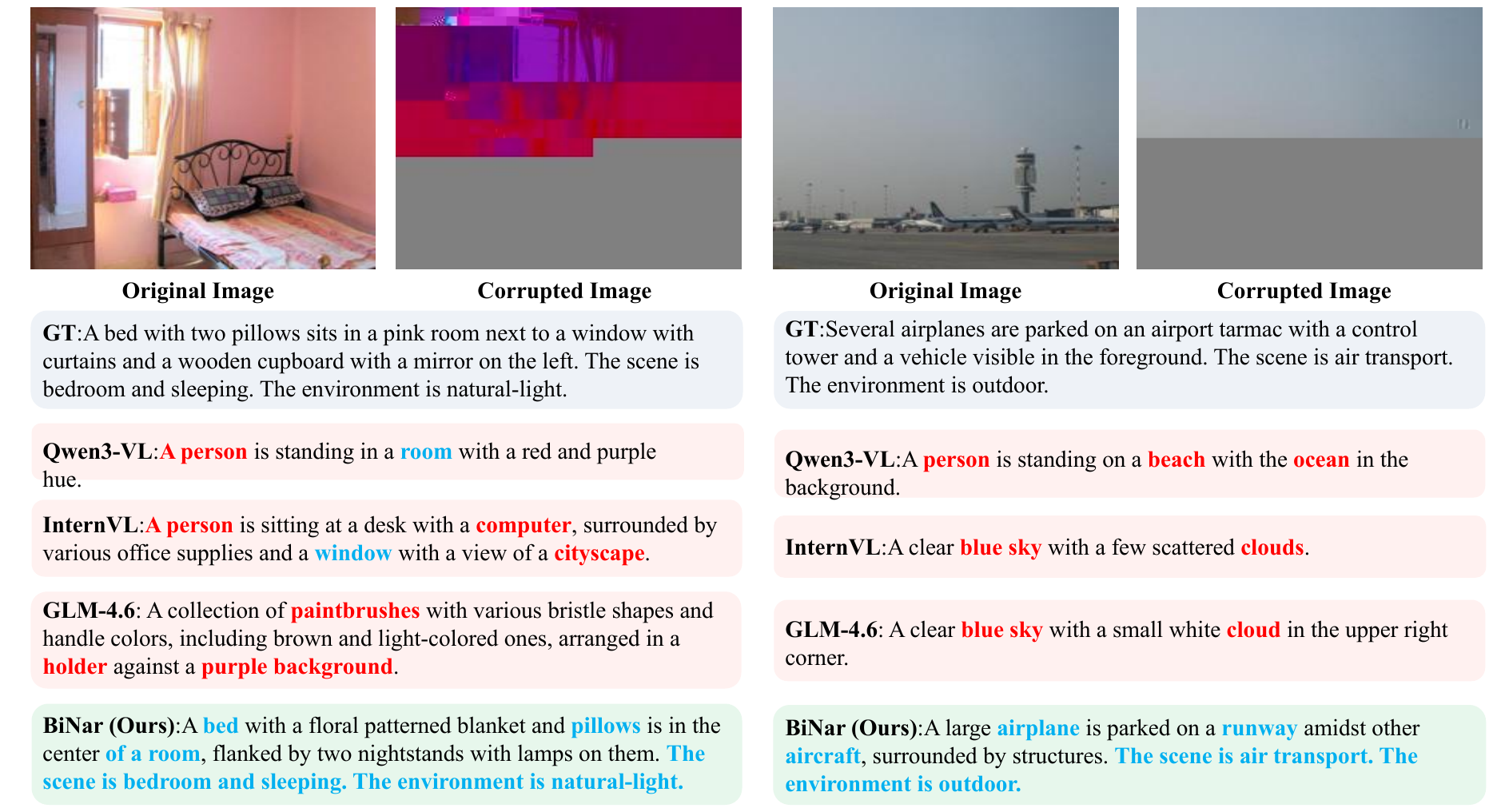}

  \captionof{figure}{Qualitative comparison of fine-grained semantic
  descriptions under bitstream corruption. BiNar preserves image content
  and scene semantics more faithfully than vision-language models relying
  on decoded corrupted images.}
  \label{fig:caption_comparison}

  \vspace{1.5mm}

  % =========================================================
  % Visual vs. Bitstream Semantics
  % =========================================================
  \includegraphics[
      width=0.99\linewidth,
      height=0.3\textheight,
      keepaspectratio
  ]{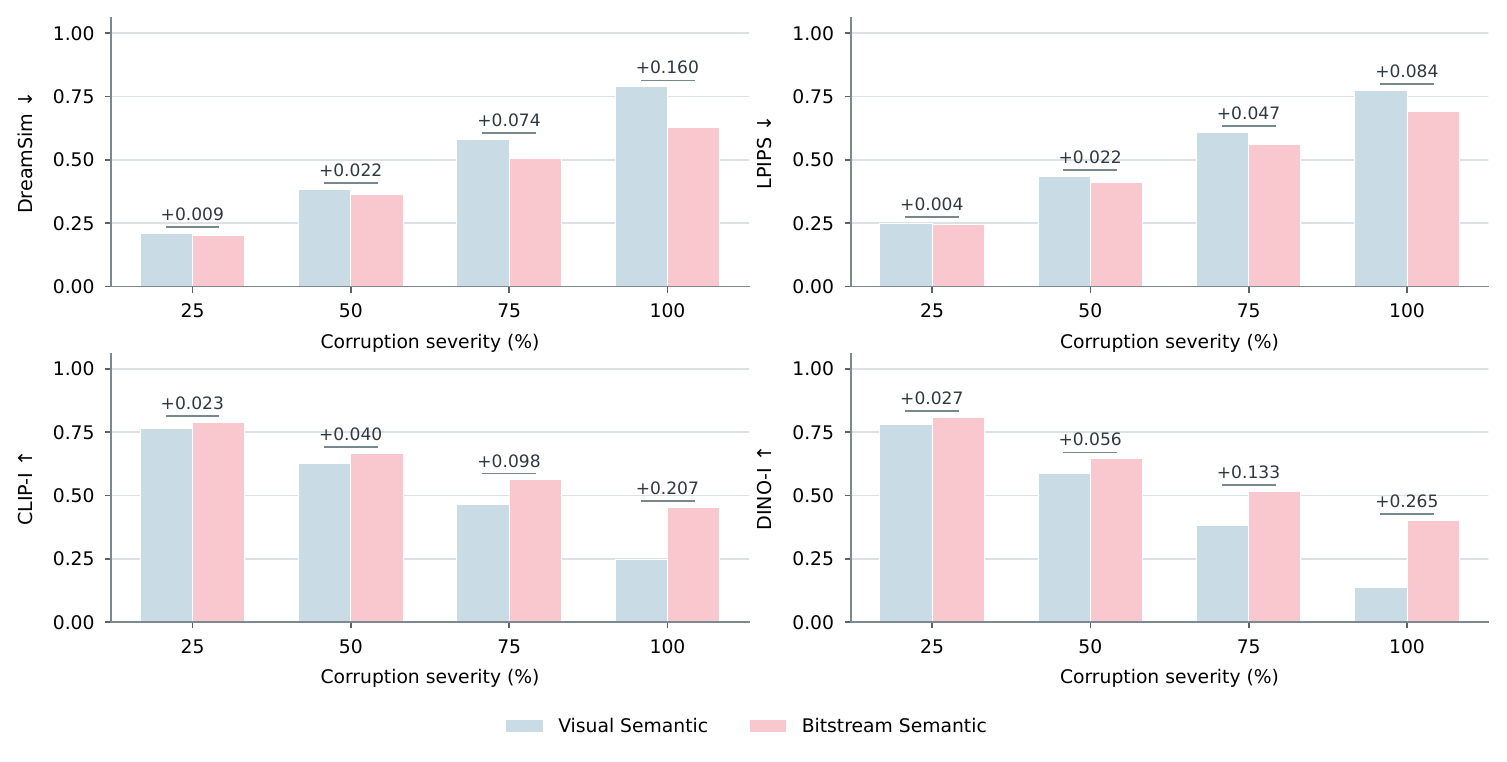}

  \captionof{figure}{Restoration quality using visual and bitstream-native
  semantics across different corruption ranges. Results are averaged over
  Bit-Flip and Byte-Loss.}
  \label{fig:semantic_source_severity}

\end{figure*}

\paragraph{Fine-grained Semantic Description Quality.}
We compare the fine-grained semantic descriptions produced by BiNar with
those generated by six representative vision-language models, including
GLM-4.6V, InternVL3, Qwen3-VL, GPT-5.6,
Gemini-3.8-Flash, and Claude Opus 4.6. The competing models infer semantics
from decoded corrupted images, whereas BiNar combines bitstream-native
semantics with useful visual cues to construct fine-grained descriptions.
As shown in Table~\ref{tab:caption_comparison}, BiNar achieves the best
performance across all reported metrics. The examples in
Fig.~\ref{fig:caption_comparison} further show that BiNar preserves image
content, Locale, and Vibe more faithfully under bitstream corruption,
demonstrating the quality and robustness of the resulting fine-grained
descriptions.

\paragraph{Robustness of Bitstream-native Semantics.}
We further evaluate the robustness of bitstream-native semantics against
semantics extracted from decoded corrupted images. The two variants use
identical corrupted inputs, localization masks, diffusion models, and sampling
settings, differing only in the semantic source. As shown in
Fig.~\ref{fig:semantic_source_severity}, both variants degrade as corruption
increases, but restoration guided by bitstream-native semantics degrades more
slowly. Consequently, the performance gap consistently widens from $25\%$ to
$100\%$ corruption across DreamSim, LPIPS, CLIP-I, and DINO-I. In particular,
the gap increases from $0.009$ to $0.160$ on DreamSim and from $0.027$ to
$0.265$ on DINO-I. These results demonstrate that bitstream-native semantics
remain more robust to increasing corruption and provide more stable restoration
guidance when semantics inferred from decoded images become unreliable.

\begin{figure*}[!t]
    \centering

    % ==================== Table ====================
    \captionof{table}{Average quantitative comparison with image restoration
    methods across all bitstream corruption configurations.}
    \label{tab:restoration_comparison}

    \resizebox{\textwidth}{!}{%
    \begin{tabular}{lcccccccc}
        \toprule
        \multirow{2}{*}{Method}
        & \multicolumn{4}{c}{Objective Metrics}
        & \multicolumn{4}{c}{Perceptual Metrics} \\
        \cmidrule(lr){2-5} \cmidrule(lr){6-9}
        & PSNR $\uparrow$ & SSIM $\uparrow$ & MAE $\downarrow$
        & VFID $\downarrow$
        & DreamSim $\downarrow$ & LPIPS $\downarrow$
        & CLIP-I $\uparrow$ & DINO-I $\uparrow$ \\
        \midrule

        Corrupted Input
        & 12.6055 & 0.5179 & 46.7493
        & 18.0492 & 0.5228 & 0.5679 & 0.5072 & 0.4676 \\

        \midrule

        LD-RPS (ICCV 2025)
        & 8.9032 & 0.1571 & 79.2295
        & 36.2306 & 0.7982 & 0.7768 & 0.2992 & 0.0734 \\

        LucidFlux (ICLR 2026)
        & 11.9684 & 0.4075 & 51.1149
        & 18.9616 & 0.4770 & 0.5294 & 0.5613 & 0.5210 \\

        SeeSR (CVPR 2024)
        & 12.5476 & 0.5012 & 47.3506
        & 16.1839 & 0.4327 & 0.4786 & 0.6018 & 0.5682 \\

        HYPIR (SIGGRAPH 2025)
        & 12.2851 & 0.4710 & 49.1705
        & 17.4295 & 0.4404 & 0.5049 & 0.5862 & 0.5566 \\

        DATPRL-IR (ICLR 2026)
        & 11.2825 & 0.4942 & 56.6913
        & 16.2785 & 0.4135 & 0.4785 & 0.6029 & 0.5920 \\

        M-GDM (CVPR 2025)
        & 12.9040 & 0.4505 & 44.3498
        & 18.9229 & 0.4977 & 0.4816 & 0.5325 & 0.4717 \\

        B2SCVR (ACM MM 2025)
        & 15.2171 & \textbf{0.5683} & 36.1204
        & 16.1131 & 0.4015 & 0.4533 & 0.5871 & 0.4886 \\

        Two-ACIR (CVPR 2023)
        & 12.6263 & 0.5156 & 46.7522
        & 32.0256 & 0.4316 & 0.4704 & 0.5803 & 0.5732 \\

        \textbf{BiNar (Ours)}
        & \textbf{15.8629}
        & 0.5198
        & \textbf{31.5975}
        & \textbf{14.8910}
        & \textbf{0.3680}
        & \textbf{0.4231}
        & \textbf{0.7210}
        & \textbf{0.7103} \\

        \bottomrule
    \end{tabular}%
    }

    \vspace{3mm}

    % ==================== Figure ====================
    \includegraphics[width=\linewidth]{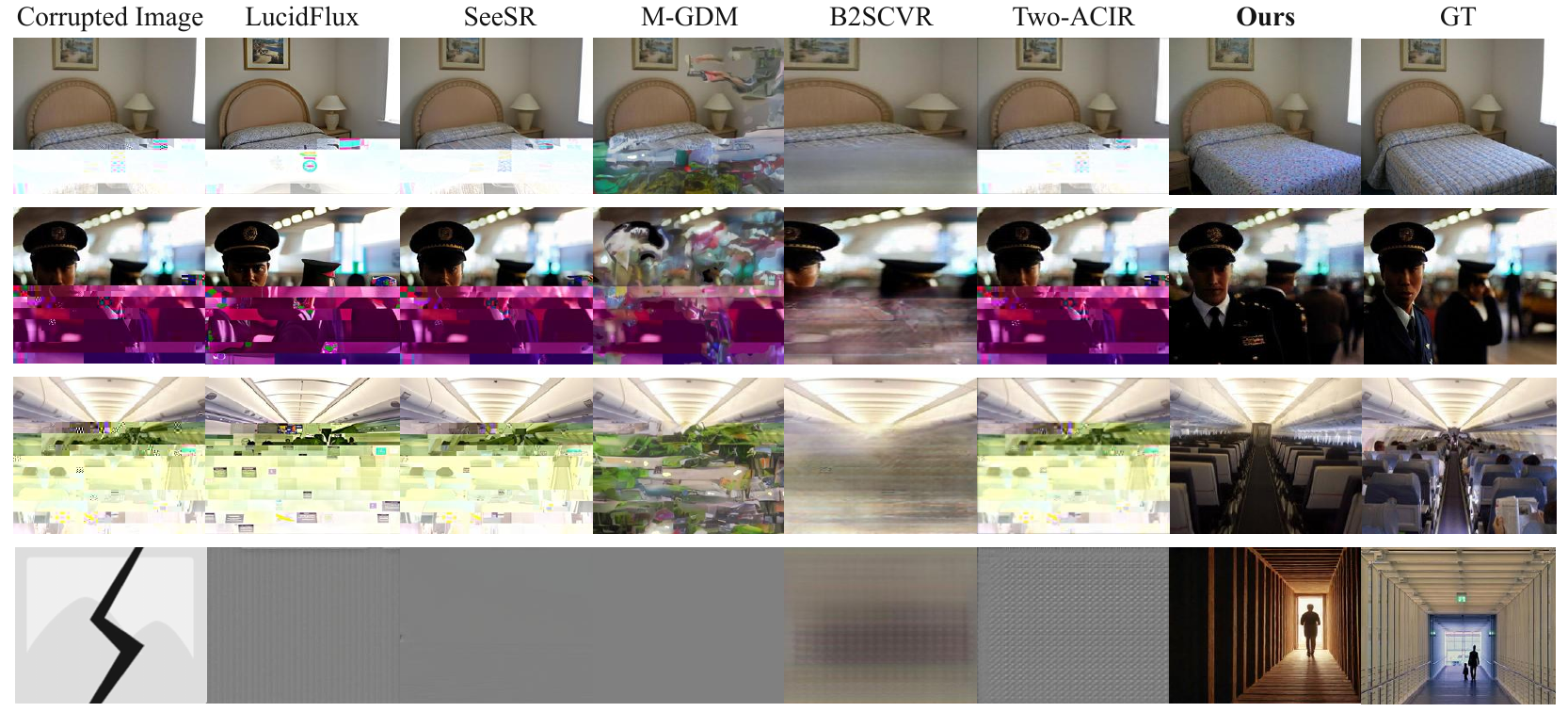}

    \captionof{figure}{Qualitative comparison with image restoration methods
    under bitstream corruption. BiNar produces more faithful restoration
    results.}
    \label{fig:restoration_comparison}

\end{figure*}

\subsubsection{Comparison with SOTA Image Restoration models}
\label{sec:restoration_comparison}

We compare BiNar with eight representative restoration methods, covering
recent generative restoration approaches and methods designed for bitstream
corruption. All methods are evaluated on the same corrupted test sets. Since
our test images contain no embedded thumbnails, methods requiring thumbnail
guidance use a downsampled version of the decoded corrupted image as auxiliary
input, following their original preprocessing without introducing clean-image
information. For video-based baselines, each corrupted image is treated as a
single-frame sequence. When a fixed temporal length is required, the same
corrupted frame is replicated along the temporal dimension, providing no
additional temporal information.

\paragraph{Quantitative Results.}
Table~\ref{tab:restoration_comparison} reports the average performance across
all bitstream corruption configurations. BiNar achieves the best results on
seven of the eight metrics, showing consistent improvements in reconstruction
fidelity, perceptual quality, and high-level content consistency. Compared with
the strongest competing results, BiNar improves PSNR from $15.2171$ to
$15.8629$ dB and reduces MAE from $36.1204$ to $31.5975$. It further reduces
VFID, DreamSim, and LPIPS to $14.8910$, $0.3680$, and $0.4231$, respectively.
These improvements indicate that the restored images are not only closer to
the clean references at the pixel level, but also exhibit better perceptual
quality.

The advantage is more pronounced on high-level representation metrics.
BiNar improves CLIP-I from $0.6029$ to $0.7210$ and DINO-I from $0.5920$ to
$0.7103$, corresponding to gains of $0.1181$ and $0.1183$, respectively.
This suggests that fine-grained bitstream-native semantics help preserve the
underlying image content when decoded pixels alone provide insufficient or
misleading information. BiNar also substantially improves over the corrupted
input itself, confirming that the gains arise from effective reconstruction
rather than simply preserving the damaged observation. Although B2SCVR obtains
the highest SSIM, BiNar performs better across all other reported metrics,
demonstrating a stronger overall balance between low-level fidelity and
perceptual consistency.

\paragraph{Qualitative Results.}
Figure~\ref{fig:restoration_comparison} provides visual comparisons under
different bitstream corruption conditions. Existing restoration methods may
retain corruption artifacts or reconstruct content that is inconsistent with
the original image when the decoded input contains substantial distortion or
missing information. In contrast, BiNar produces more coherent structures and
recovers image content that is better aligned with the clean reference. These
visual results are consistent with the quantitative improvements, particularly
on LPIPS, CLIP-I, and DINO-I, and further demonstrate the benefit of combining
fine-grained bitstream-native semantics with spatial corruption localization.

\begin{table*}[!t]
\centering

% ==================== BSeM Ablation ====================
\caption{Ablation on Plot--Scene decomposition in Bitstream Semantic Modeling (BSeM).}
\label{tab:structured_semantic_ablation}
\scriptsize
\resizebox{\textwidth}{!}{%
\begin{tabular}{lcccccc}
\toprule
Method
& BLEU-1
& BLEU-4
& ROUGE-L
& CIDEr
& Token-F1
& BERTScore-F1 \\
\midrule
Unified Semantic Generation
& 0.3026
& 0.0686
& 0.2901
& 0.0426
& 0.3335
& 0.9680 \\
\textbf{Plot--Scene Decomposition}
& \textbf{0.3743}
& \textbf{0.1611}
& \textbf{0.3682}
& \textbf{0.1169}
& \textbf{0.4200}
& \textbf{0.9718} \\
\bottomrule
\end{tabular}%
}

\vspace{-1mm}

% ==================== CMSA Ablation ====================
\caption{Ablation on Corruption-adaptive Multimodal Semantic Aggregation (CMSA).}
\label{tab:cmsa_ablation}
\scriptsize
\resizebox{\textwidth}{!}{%
\begin{tabular}{lcccccc}
\toprule
Method
& BLEU-1
& BLEU-4
& ROUGE-L
& CIDEr
& Token-F1
& BERTScore-F1 \\
\midrule
Visual Only
& 0.2395
& 0.1295
& 0.3032
& 0.2634
& 0.3521
& 0.6354 \\
Bitstream Only
& 0.3919
& 0.1826
& 0.4187
& 0.1091
& 0.4771
& 0.9716 \\
\textbf{CMSA}
& \textbf{0.4541}
& \textbf{0.2248}
& \textbf{0.4367}
& \textbf{0.2934}
& \textbf{0.5028}
& \textbf{0.9776} \\
\bottomrule
\end{tabular}%
}

\end{table*}

\subsection{Validation Experiments}
\label{sec:validation}
We conduct controlled experiments to validate three key designs of BiNar,
including the Plot--Scene decomposition in BSeM, corruption-adaptive
multimodal semantic aggregation, and blind corruption localization.

\paragraph{Ablation on Plot--Scene Decomposition in BSeM.}
We examine the structured semantic modeling design in BSeM by comparing it
with unified semantic generation. The unified variant directly generates a
complete semantic description, while BSeM decomposes the bitstream semantics
into Plot and Scene, with Scene further characterized by Locale and Vibe. As
shown in Table~\ref{tab:structured_semantic_ablation}, the proposed
decomposition consistently improves all semantic description metrics,
demonstrating the effectiveness of structured semantic modeling in BSeM.

\paragraph{Ablation on Corruption-adaptive Multimodal Semantic Aggregation.}
We compare visual-only, bitstream-only, and the full CMSA. As shown in
Table~\ref{tab:cmsa_ablation}, the bitstream-only variant outperforms the
visual-only variant on most metrics, while CMSA further improves semantic
quality by incorporating complementary visual cues when they remain
informative.

\paragraph{Ablation on Blind Corruption Localization.}
We compare BCoL with a generic SAM-based localization baseline. BCoL keeps
the pretrained visual encoder frozen and trains only a lightweight corruption
localization head. It improves mIoU from $0.7957$ to $0.9925$ and Dice-F1
from $0.8765$ to $0.9962$, demonstrating substantially more accurate
localization of bitstream-induced corruption.

\section{Conclusion}

In this work, we presented Bitstream Native Restorer (BiNar), a
generative framework for bitstream-corrupted image restoration that exploits
fine-grained bitstream-native semantics for image reconstruction. BiNar uses
Bitstream Semantic Modeling (BSeM) to extract semantic information
directly from corrupted bitstreams without relying on pixel decoding.
Corruption-adaptive Multimodal Semantic Aggregation (CMSA) further
aggregates useful visual cues with the bitstream-derived semantics, while
Blind Corruption Localization (BCoL) identifies regions requiring
recovery. The resulting semantic descriptions and spatial prompts jointly
guide diffusion-based restoration.
Extensive experiments demonstrate that bitstream-native semantics remain more
informative than pixel-derived semantics under bitstream corruption and provide
effective guidance for restoring both visual quality and semantic content.
BiNar consistently achieves strong restoration performance across diverse
corruption conditions, while ablation studies validate the effectiveness of
its key components. These results highlight corrupted bitstreams as a valuable
native source of semantic information for robust image restoration.

\bibliography{iclr2027_conference}
\bibliographystyle{iclr2027_conference}
\end{document}